\documentclass[10pt,twocolumn,letterpaper]{article}

\usepackage{iccv}
\usepackage{times}
\usepackage{epsfig}
\usepackage{graphicx}
\usepackage{amsmath}
\usepackage{amssymb}
\usepackage{capt-of}
\usepackage{comment}
\usepackage{booktabs}
\usepackage{multirow,tabularx}
\usepackage{float}
\usepackage[dvipsnames]{xcolor}
\usepackage{url}
\newcommand{\mybullet}{\vspace{0.05cm}\noindent $\bullet$\ }
\newcommand{\mybulletend}{\vspace{0.05cm}}

\newcommand{\mysubsubsection}[1]{\vspace{0.1cm} \noindent {\bf #1}:}

\usepackage[breaklinks=true,bookmarks=false]{hyperref}
\iccvfinalcopy 

\def\iccvPaperID{1233} 

\ificcvfinal\fi

\title{Floor-SP: Inverse CAD for Floorplans by\\ Sequential Room-wise Shortest Path}

\author{Jiacheng Chen$^1$ \qquad Chen Liu$^2$ \qquad Jiaye Wu$^2$ \qquad Yasutaka Furukawa$^1$ \\
$^1$Simon Fraser University \qquad $^2$Washington University in St. Louis \\
{\tt\small \{jca348,furukawa\}@sfu.ca} \qquad {\tt\small \{chenliu,jiaye.wu\}@wustl.edu}}

\begin{document}
\twocolumn[{
\maketitle
\vspace{-2em}
\centerline{
\includegraphics[width=\textwidth]{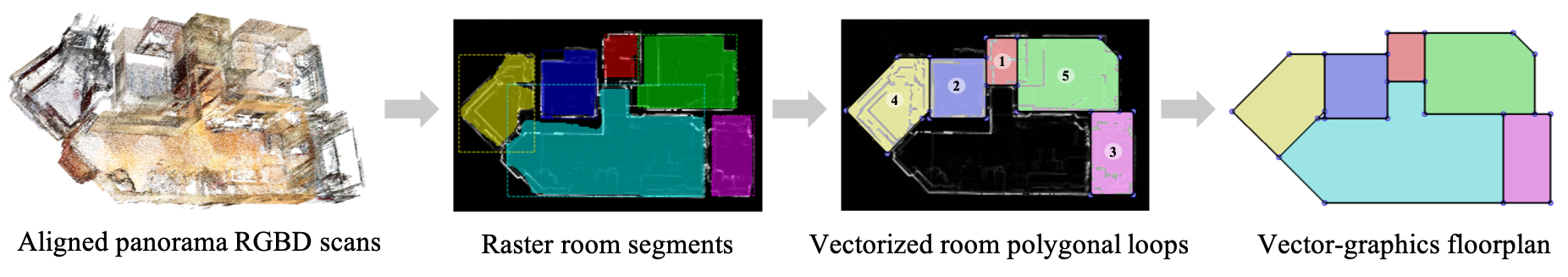}
}
\captionof{figure}
{
The proposed system, dubbed {\it Floor-SP}, takes aligned panorama RGBD scans as input, finds room segments, solves an optimization problem to reconstruct a floorplan graph as multiple polygonal loops (one for each room), and merges them into a 2D graph via simple post-processing heuristics. The optimization is the technical contribution of the paper, which employs the room-wise coordinate descent strategy and sequentially solves shortest path problems to optimize the room structure.
}
\label{fig:teaser}
\vspace{1em}
}]

\begin{abstract}
\vspace{-1em}
This paper proposes a new approach for automated floorplan reconstruction from RGBD scans, a major milestone in indoor mapping research. The approach, dubbed Floor-SP, formulates a novel optimization problem, where room-wise coordinate descent sequentially solves shortest path problems to optimize the floorplan graph structure.
%
The objective function consists of data terms guided by deep neural networks, consistency terms encouraging adjacent rooms to share corners and walls, 
and
the model complexity term.
%
The approach does not require corner/edge primitive extraction unlike most other methods.
%
We have evaluated our system on production-quality RGBD scans of 527 apartments or houses,
including many units with non-Manhattan structures.
Qualitative and quantitative evaluations demonstrate a significant performance boost over the current state-of-the-art.
Please refer to our project website \url{http://jcchen.me/floor-sp/} for code and data.
\end{abstract}

\section{Introduction}

Architectural floorplans play a crucial role in designing, understanding, and remodeling indoor spaces. Automated floorplan reconstruction from raw sensor data is a major milestone in indoor mapping research.
%
The core technical challenge lies in the inference of wall graph structure, whose topology is unknown and varies per example.

Computer Vision has made remarkable progress in the task of graph inference, for instance, human pose estimation~\cite{openpose} and hand tracking~\cite{yuan2017bighand2}. Unfortunately, the success has been limited to the cases of fixed known topology (e.g., a human has two arms). Inference of graph structure with unknown varying topology
is still an open problem.

A popular approach to graph reconstruction is {\it primitive detection and selection}~\cite{manhattan_world_stereo,reconstructing_world_museum,monszpart2015rapter}, for example, detecting corners, selecting subsets of corners to form edges, and selecting subsets of edges to form regions.
The major problem of this bottom-up process is that it cannot recover from a single false-negative in an earlier stage (i.e., a missing primitive). The task becomes increasingly more difficult as the primitive space grows exponentially with their degrees of freedom, especially for non-Manhattan scenes which most existing methods do not handle~\cite{manhattan_world_stereo,cabral2014piecewise,liu2017raster,liu2018floornet}. 
%

This paper seeks to make a breakthrough in the domain of floorplan reconstruction with three key ideas.

\mybullet
First, we start from room segmentation via instance semantic segmentation technique (we use Mask-RCNN ~\cite{maskrcnn}). The room segmentation reduces the floorplan graph inference into the reconstruction of multiple polygonal loops, one for each room. This reduction allows us to formulate floorplan reconstruction as sound energy optimization over multiple loops guided by room proposals.

\mybullet
Second, we employ {\it room-wise coordinate descent} strategy in optimizing the objective function. By exploiting the fact that the room topology is a simple loop, our formulation finds the (near-)optimal graph structure by solving a shortest path problem for each room one by one
sequentially, while enforcing consistency with the other rooms.

\mybullet
Third, we utilize deep neural networks in evaluating the data terms of the optimization problem, measuring the discrepancy against the input sensor data. The data term is combined with the ad-hoc 1)
consistency term, encouraging adjacent rooms to share corners and walls at the room boundaries, and 2) model complexity term, penalizing the number of corners in the graph. 

\mybulletend
We have evaluated the proposed approach on production-quality RGBD scans of 527 apartments or houses, a few times larger than the current largest database~\cite{liu2018floornet}. Our approach makes significant improvements over the current state-of-the-art~\cite{liu2018floornet}. We refer to our project website \url{http://jcchen.me/floor-sp/} for code and data.


\section{Related Works}
We discuss related work in two domains: graph reconstruction and indoor scan datasets.

\mysubsubsection{Graph reconstruction}
Graph structure inference has been a popular field of study in Computer Vision, for instance, inferring a human body pose~\cite{openpose} or the semantic relationships of categories~\cite{Hu2016LearningSI,xu2017scene}.
In these problems, the graph topology is defined over the label space, common to all the instances (e.g., a {\it head} is always connected to a {\it body}).
We here focus on graph inference problems in the context of reconstruction, where the topology varies per instance.


Room layout estimation infers a graph of architectural feature lines from a single image, where nodes are room corners and edges are wall boundaries. Most approaches assume a 3D box-room to limit the topological variations in the room layouts visible in 2D images~\cite{hedau2009recovering,schwing2012efficient,lee2017roomnet,chao2013layout}.
%
For a room beyond a box shape, 
Dynamic Programming (DP) was applied to search for an optimal room structure~\cite{flint2010dynamic,flint2011manhattan}. DP was similarly used to solve for floorplans by limiting their topology to be a loop~\cite{cabral2014piecewise}.
%

%
%


Bottom-up processing is a popular approach for graph reconstruction, where low-level primitives such as corners are detected, which are then selected to form higher-level primitives such as edges or regions.
%
DNN-based junction detector was proposed for floorplan image vectorization~\cite{liu2017raster}, where a junction indicates incident edge directions in the Manhattan frame. The junction information is utilized in inferring the edges by integer programming (IP). Similarly, Huang~\etal~\cite{wireframe_cvpr18} uses DNN to detect junctions represented by a set of incident edge directions, and infer edges by heuristics for single-image wireframe reconstruction of man-made scenes.

While many previous works utilize RGBD scans/point clouds for high-quality indoor reconstruction~\cite{ikehata2015structured, Li2016ManhattanWorldRECON, Nan2017PolyFitPS, liu2018floornet}, FloorNet~\cite{liu2018floornet} is the current state-of-the-art for floorplan reconstruction task tested on large-scale indoor benchmarks. FloorNet combines DNN and IP in a bottom-up process but it has three major failure modes. 
First, as in any bottom-up process, missing corners in the detection phase automatically lead to missing walls and rooms in the final model. Second, false candidate primitives could lead to the reconstruction of extraneous walls and rooms. Third, to enable the usage of powerful IP, FloorNet needs to restrict the solution space to Manhattan scenes.

Structured indoor modeling by Ikehata~\etal~\cite{ikehata2015structured} is the source of inspiration for our work, which starts by room segmentation then solves shortest path problems to reconstruct room shapes followed by room merging and room addition. While their system is a sequence of heuristics for indoor modeling, our approach formulates a sound energy minimization problem to recover the floorplan structure.



\mysubsubsection{Indoor scan datasets}
Affordable depth sensing hardware enables researchers to build many indoor scan datasets. The ETH3D dataset contains 16 indoor scans for multi-view stereo~\cite{schops2017multi}.
The ScanNet dataset~\cite{dai2017scannet} and the SceneNN dataset~\cite{hua2016scenenn} capture a variety of indoor scenes. However, most of their scans contain only one or two rooms, not suitable for the floorplan reconstruction problem. Matterport3D~\cite{chang2017matterport3d} builds high-quality panorama RGBD image sets for 90 luxurious houses. 2D-3D-S dataset~\cite{armeni20163d} provides 6 large-scale indoor scans of office spaces by using the same Matterport system. Lastly, a large-scale synthetic dataset, SUNCG~\cite{song2016semantic}, offers a variety of indoor scenes.

For the floorplan reconstruction task, FloorNet~\cite{liu2018floornet} provides the benchmark with full floorplan annotations and the corresponding RGBD videos from smartphones for 155 residential units. This paper utilizes production-quality panorama RGBD scans for 527 houses or apartments with floorplan annotations.

\section{Floor-SP: System Overview} \label{sect:formulation}

\begin{figure*}[tb]
    \centering
    \includegraphics[width=\textwidth]{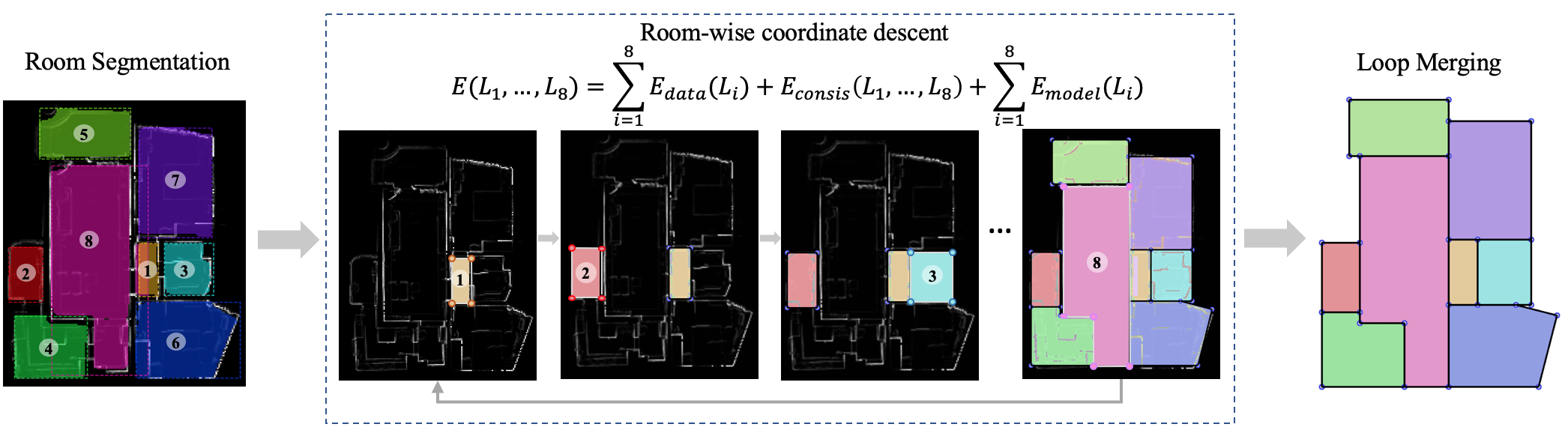}
    \caption{System overview: (Left) Mask-RCNN finds room segments (raster) from a top-down projection image consisting of point density and mean surface normal, allowing us to reconstruct a floorplan as multiple room loops. (Middle) Room-wise coordinate descent optimizes vectorzied room structures one by one by minimizing the sum of data, consistency, and model complexity terms. (Right) Simple graph merging operations combine loops into a floorplan graph structure.
    }
    \label{fig:system}
\vspace{-1em}
\end{figure*}

Floor-SP turns aligned panorama RGBD images into a floorplan graph in three phases: room segmentation, room-aware floorplan reconstruction, and loop merging (See Fig.~\ref{fig:system}). This section provides the system overview with minimal details. The aligned panorama RGBD scans are first converted into 2D point-density/normal map, which is the input to Floor-SP.
Unlike FloorNet~\cite{liu2018floornet}, we focus on the wall structures, where doors/windows, icons, and room semantics can be added given proper wall structures.

\mysubsubsection{Room segmentation}
The input panorama scans are converted into a 4-channel $256\times 256$ point-density/normal map in a top-down view 
(See Sect.~\ref{sect:details}).
We utilize instance semantic segmentation technique (Mask R-CNN~\cite{maskrcnn}) to find room segments given the 4-channel image. 
The room segments set up a good foundation for floorplan reconstruction by providing room proposals with rough shape, but they are still far away from a good floorplan graph because 1) Mask R-CNN segment has a raster representation (i.e., unknown number and placement of corners); and 2) Walls are not consistently shared across rooms.
%

\mysubsubsection{Room-aware floorplan reconstruction}
Given a set of room segments and the input point-density/normal map, we formulate an optimization problem that reconstructs a floorplan graph as multiple polygonal loops, one for each room.
Deep neural networks derive data terms in the objective.
We propose a novel room-wise coordinate descent algorithm that directly optimizes the number and placement of corners by sequentially solving shortest-path problems.

\mysubsubsection{Loop merging}
Simple graph merging operations combine multiple polygonal loops into a final floorplan graph.

\mybulletend
Room-aware floorplan reconstruction is the technical core of the paper, where Sect.~\ref{sect:problem} defines the problem formulation, and Sect.~\ref{sect:algorithm} presents the optimization algorithm. Room segmentation and loop merging are based on existing techniques, where Sect.~\ref{sect:details} provides their algorithmic details and the remaining system specifications.
\section{Room-aware floorplan reconstruction} \label{sect:problem}
The room segmentation ($R_i$) from Mask R-CNN allows us to reduce the floorplan graph inference into the reconstruction of multiple loops ($L_i$), one for each room. $L_i$ is defined as a sequence of pixels at integer coordinates forming a polygonal curve with a loop topology.
Our problem is to minimize the following objective with respect to 
the set of polygonal loops $\mathcal{L}$:
\begin{eqnarray*}
\sum_{L_i \in \mathcal{L}} E_{data}(L_i) + E_{consis}\left(\mathcal{L}\right) + \sum_{L_i \in \mathcal{L}} E_{model}(L_i),
\end{eqnarray*}subject to $L_i$ being a loop containing \mbox{$R_i$} inside. Note that a room has an arbitrary number of corners (i.e., degrees of freedom), which must be optimized by an algorithm.

\mysubsubsection{Data term}
$E_{data}$ is a room-wise unary potential, measuring the discrepancy with the input sensor data over the set of pixels along each loop.
%
\begin{eqnarray*}
E_{data}(L_i) &=& \sum_{p\in \mathbb{C}(L_i)} \lambda_1 E_{data}^{\mathcal{C}}(p) +\\
&& \sum_{p\in \mathbb{E}(L_i)} \left[\lambda_2 E_{data}^{\mathcal{E}}(p) +
 \lambda_3 E_{data}^{\mathcal{I}}(p)\right].
\end{eqnarray*}

\begin{figure}
    \centering
    \includegraphics[width=\columnwidth]{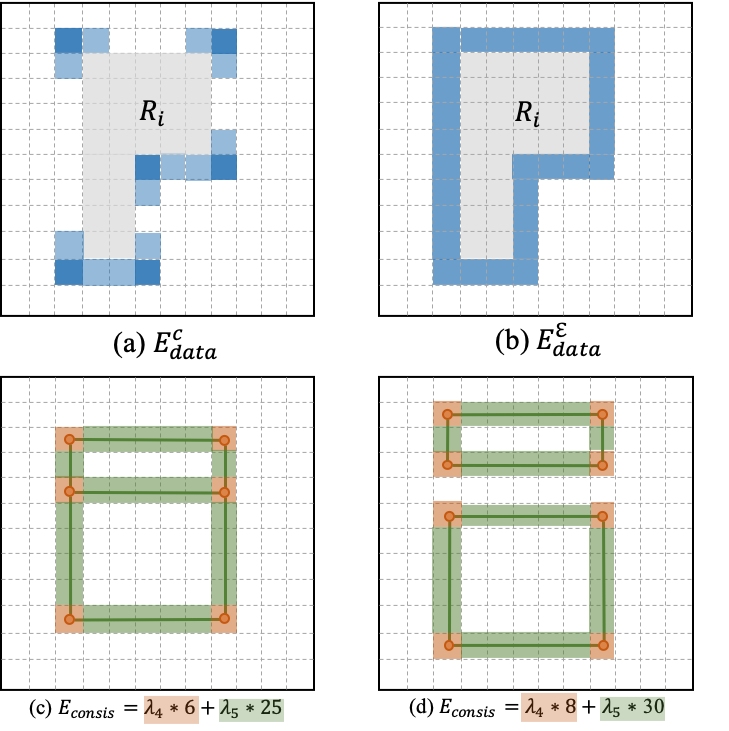}
    \caption{Illustration of data and consistency terms.
    $E_{data}^{\mathcal{C}}$ and $E_{data}^{\mathcal{E}}$ are defined based on corner and edge likelihood maps. Blue pixels indicate lower costs in these toy examples. $E_{consis}$ counts the number of pixels used by room corners and room edges. When neighboring rooms share corners and edges as shown in (c), $E_{consis}$ goes down.}
    \label{fig:energy}
    \vspace{-1em}
\end{figure}

\mybullet
$E_{data}^{\mathcal{C}}(p)$ is the penalty of placing a corner at pixel {\it p} (see Fig.\ref{fig:energy}a), and hence, summed over all the corner pixels $\mathbb{C}(L_i)$ on $L_i$. The penalty is defined as one minus the pixel-wise corner likelihood. We estimate the corner likelihood map from the input point-density/normal map using Dilated Residual Networks (DRN)~\cite{Yu2017DilatedRN}.

\mybullet
$E_{data}^{\mathcal{E}}(p)$ is the penalty of placing an edge over a pixel {\it p}. The term is defined as one minus the pixel-wise edge likelihood (see Fig.~\ref{fig:energy}b), summed over all the edge pixels $\mathbb{E}(L_i)$ along $L_i$. We use Bresenham's line algorithm to obtain edge pixels given corners. The same DRN estimates the edge likelihood from the input point-density/normal map.

\mybullet
$E_{data}^{\mathcal{I}}(p)$ is also the penalty summed over the edge pixels, which enforces $L_i$ not to pass through the room segment $R_i$. The term is a large constant if a pixel belongs to any of the room segments and 0 otherwise.

%
%
%

\mysubsubsection{Consistency term}
$E_{consis}$ is a room-wise higher-order potential, encouraging loops to be consistent at the room boundaries (i.e., sharing corners and edges). We define the penalty to be the number of pixels that are {\it used by} the corners (or edges) of all the loops together.
For instance, if two corners are close to each other, this term suggests to move them to the same pixel so that penalty is imposed only once:
%
\begin{eqnarray*}
E_{consis}(\mathcal{L}) = \sum_{p} \left[ \lambda_4{\bf 1}_{\mathcal{C}}(p, \mathcal{L}) \right] + \sum_{p} \left[\lambda_5{\bf 1}_{\mathcal{E}}(p, \mathcal{L}) \right]
\end{eqnarray*}
The first term ${\bf 1}_{\mathcal{C}}(p, \mathcal{L})$ is an indicator function, which becomes 1 if a pixel (p) is a corner of at least one loop. Similarly, the second term is an indicator function for edges. See Fig.~\ref{fig:energy} for the illustration over toy examples.
%

\mysubsubsection{Model complexity term} $E_{model}$ is the model complexity penalty, counting the number of corners in our loops, preferring compact shapes.
\begin{eqnarray*}
E_{model} (L_i) = \lambda_6 \text{\{\# of corners in $L_i$\}}.
\end{eqnarray*} 
 
\mybulletend
$\lambda_{?}$ are scalars defining the relative weights of the penalty terms. We found our system robust to these parameters and use the following setting throughout our experiments: $\lambda_1=0.2, \lambda_2=0.2, \lambda_3=100.0, \lambda_4=0.2, \lambda_5=0.1, \lambda_6=1.0$.
\section{Sequential room-wise shortest path} 
\label{sect:algorithm}

The inspiration of our optimization strategy comes from a prior work, which solves a shortest path problem and reconstructs a floorplan as a loop~\cite{cabral2014piecewise}. This formulation considers every pixel as a node of a graph, encodes objectives into edge weights, and finds the shortest path as a loop. 

Our problem solves for multiple loops over multiple rooms. We devise {\it room-wise coordinate descent strategy} that optimizes room structures one by one sequentially by reducing a room-wise coordinate descent step into a shortest path problem. 
While the algorithm is robust to the processing order, we visit rooms in increasing order of their areas (i.e. smaller rooms are handled first) so that we get fixed results given the same input. The optimization runs for two rounds in our experiments. 

This section explains 1) Shortest path problem reduction; 2) Containment constraint satisfaction; and 3) Two approximation methods for speed-boost.

\mysubsubsection{Shortest path problem reduction}
The reduction process is straightforward, as our cost function is the summation of pixel-wise penalties and the number of corners.
Without loss of generality, suppose we are optimizing $L_1$ while fixing the other loops. Our optimization problem is equivalent to solving a shortest path problem for $R_1$ with the following weight definition for each edge $(e)$ (See the supplementary document for the derivation):
\begin{eqnarray*}
 \sum_{p\in \mathbb{C}(e)} \frac{\lambda_1}{2} E_{data}^{\mathcal{C}}(p) &+& \\
 \sum_{p\in \mathbb{E}(e)} \left[\lambda_2 E_{data}^{\mathcal{E}}(p) +
 \lambda_3 E_{data}^{\mathcal{I}}(p)\right] &+& \\
 \sum_{p \in \mathbb{C}(e)} \lambda_4 (1 - {\bf 1}_{\mathcal{C}}(p, \mathcal{L} \setminus \{L_1\})) &+& \\
\sum_{p \in \mathbb{E}(e)} \lambda_5(1 - {\bf 1}_{\mathcal{E}}(p, \mathcal{L} \setminus \{L_1\})) &+& \lambda_6.
\end{eqnarray*}
With abuse of notation, $\mathbb{C}(e)$ denotes the two pixels at the end-points of $e$, $\mathbb{E}(e)$ denotes the set of pixels along $e$ obtained by Bresenham's line algorithm, and $\mathcal{L} \setminus \{L_1\}$ denotes the set of loops excluding $L_1$.

\mysubsubsection{Containment constraint satisfaction}
Shortest path is a powerful formulation that searches for the optimal number and placement of corners with one caveat: An additional constraint is necessary to avoid a trivial solution (i.e., an empty loop).
We use a heuristic similar in spirit to the prior work~\cite{cabral2014piecewise} to implement this constraint: ``$L_i$ contains (or goes around) $R_i$''. We refer the details to the supplementary document and here summarize the process.
\begin{figure}
    \centering
    \includegraphics[width=\columnwidth]{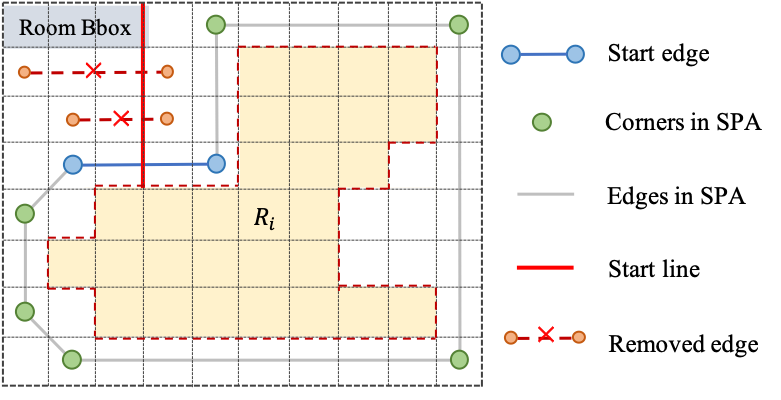}
    \caption{We solve a shortest path problem for each room, where cost functions are encoded into edge weights. In order to avoid a trivial solution (i.e., an empty graph) and enforce the path to go around the rough room segment ($R_i$), we first identify a start-edge that is a part of a room shape with high-confidence. Next, we draw a (red) start-line perpendicularly to split the domain. We prohibit crossing the start-line, assign a very high penalty for going through $R_i$, then solve for a shortest path that starts and ends at the two end-points of the start-edge.
    %
    }
    \label{fig:startline}
\vspace{-1em}
\end{figure}

First, we find corner candidates from the same corner likelihood map used for the data term (see Fig.~\ref{fig:startline}). Second, we look at the edge likelihood map to identify a good pair of corners forming the {\it start-edge} of the loop. 
Third, we draw a {\it start-line} that starts from the room mask $(R_i)$ and passes through the start-edge perpendicularly at its middle point. Lastly, we remove all the edges that intersect with the start-line 
to ensure that the path must go around $R_i$.

Note that fixing the start-edge to be part of the loop breaks the local optimality of our coordinate descent step, but works well in practice as it is not difficult to identify one wall segment with high confidence.

\mysubsubsection{Bounding box approximation}
We make an approximation in pruning nodes and edges to reduce the computational expenses of the shortest path algorithm (SPA).
We restrict the domain of SPA, as it is wasteful to run it over an entire image domain to reconstruct one room.
Given a room mask $R_i$, we apply the binary dilation 10 times to expand the mask and find its axis-aligned bounding box with a 5-pixel margin, in which we solve SPA.

\mysubsubsection{Dominant direction approximation}
Floor-SP goes beyond the conventional Manhattan assumption by allowing multiple Manhattan frames per room.
We train the same DRN architecture 
to estimate the wall direction likelihoods in an increment of 10 degrees at every pixel. We perform a simple statistical analysis to extract four Manhattan frames (i.e., eight directions) globally , then assign its subset to each room.
We allow edges only along the selected dominant directions with some tolerance on discretization errors
(See the supplementary document for details).

\section{System Details}\label{sect:details}

\mysubsubsection{Input processing}
Given a set of panorama RGBD scans where the Z axis is aligned with the gravity direction, we compute the tight axis-aligned bounding box of the points on the horizontal plane. We expand the rectangle by $2.5\%$ in each of the four directions, apply non-uniform scaling into a $256\times 256$ pixel grid, and compute the point density and normal in each pixel. The point density is the number of 3D points that fall inside the pixel, which we linearly re-scale to [0.0, 1.0] so that the highest density becomes $1.0$. The point normal is the average surface normal vector of the 3D points associated with the pixel. 

\mysubsubsection{Room segmentation}
We use the publicly available Mask R-CNN implementation~\cite{pytorch_maskrcnn} with the default hyper-parameters except that we lower the detection threshold from 0.7 to 0.2.
Given a segment from Mask R-CNN, we apply the binary erosion operation for 2 iterations 
with 8-connected neighborhood
to obtain room segments ($R_i$).

\mysubsubsection{Room-aware floorplan reconstruction}
To estimate pixel-wise likelihoods for corner, edge, and edge direction, we use the official implementation of Dilated Residual Networks~\cite{Yu2017DilatedRN}, which produces $32\times 32$ feature maps. In order to produce an output in the same resolutions as the input, we add 3 extra layers of residual blocks~\cite{resnet} with transposed convolution of stride 2 to reach the resolution of $256\times 256$.
%
For the corner likelihood supervision, we render each ground truth corner as a $7 \times 7$ disk.
For the edge likelihood and wall-direction supervision, we draw the edge mask and direction information with a width of $5$ pixels.
The loss 
is binary cross entropy and the learning rate is 1e-4. Dijkstra's algorithm solves the shortest path problem.



\mysubsubsection{Loop merging} 
We use simple graph merging operations to convert room loops into the final floorplan graph structure. 
More concretely, we denote a contiguous set of colinear line segments as a {\it segment group}. 
We repeatedly identify a pair of parallel segment groups within 5 pixels and snap them into a new segment group at the middle point while merging corners. After applying the edge merging to all compatible pairs, we merge corners that are within 3 pixels.


\section{Experiments}
\label{sect:exp}
\begin{figure*}[tb]
    \centering
    \includegraphics[width=\textwidth]{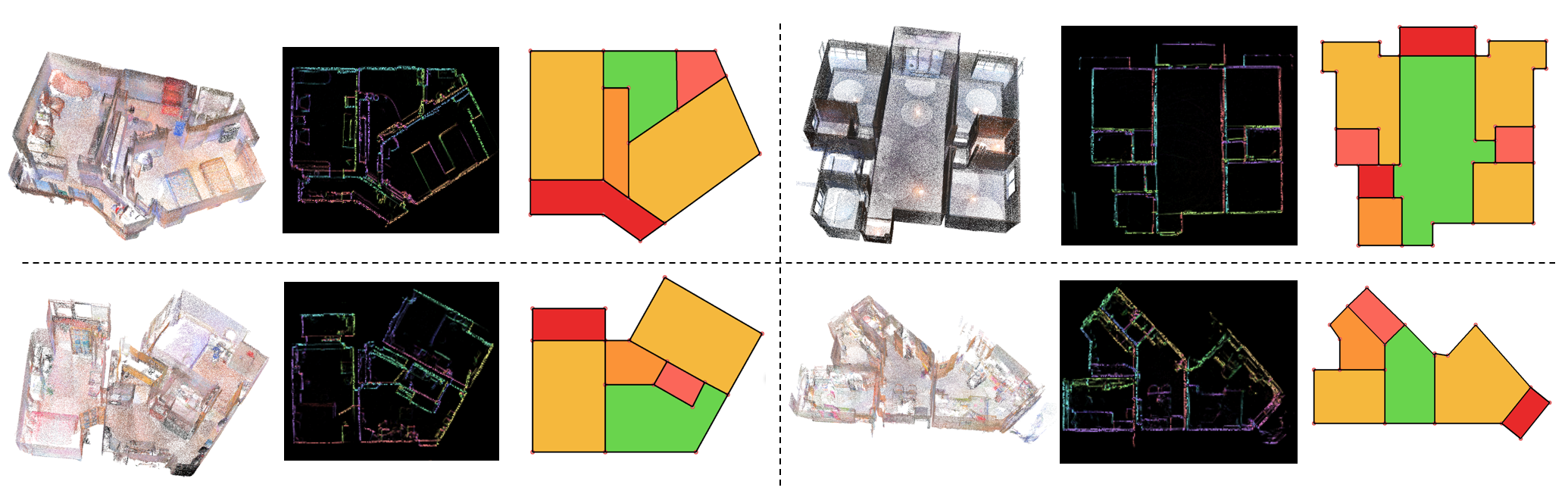}
    \caption{Our dataset offers production-level panorama RGBD scans for 527 houses/apartments.
    We convert each scan into a point density/normal map from a top-down view, which is the input to our system.
    %
    We annotated floorplan structure as a 2D polygonal graph. 
    %
    Note that for visualizing point-density/normal maps (the middle column), the intensity encodes the point density, and the hue/saturation encodes the 2D horizontal component of the mean surface normal.
}
\label{fig:dataset}
\end{figure*}

We have evaluated the proposed system on 527 sets of aligned panorama RGBD scans. The average numbers of 1) input 3D points for the point-density/normal image, 2) corners in the annotations, 3) wall segments in the annotations, and 4) rooms in the annotations are 432,552, 28.87, 35.88, and 7.73, respectively. Out of 4072 rooms, 489 rooms do not follow the primary Manhattan structure of the unit. Fig.~\ref{fig:dataset} shows four examples from our dataset. 

527 units are split into 433 and 94 for training and testing, respectively. We make the test set more challenging on purpose for evaluations:
48 out of 94 testing units contain challenging non-Manhattan structure, and 199 out of 667 testing rooms follow non-Manhattan geometry.

We have implemented the proposed system in Python while using PyTorch as the DNN library. We have used a workstation equipped with an NVIDIA 1080Ti with 12GB GPU memory.
We trained the Mask-RCNN for 70 epochs with a batch size of 1, and the DRNs for 35 epochs with a batch size of 4.
The training of each DNN model takes at most a day.
At test time, it takes about 5 minutes to process one apartment/house. The bottleneck is the construction of the graph for the shortest path problem (a CPU-intensive).




\begin{figure*}[h!]
    \centering
    \includegraphics[width=\textwidth]{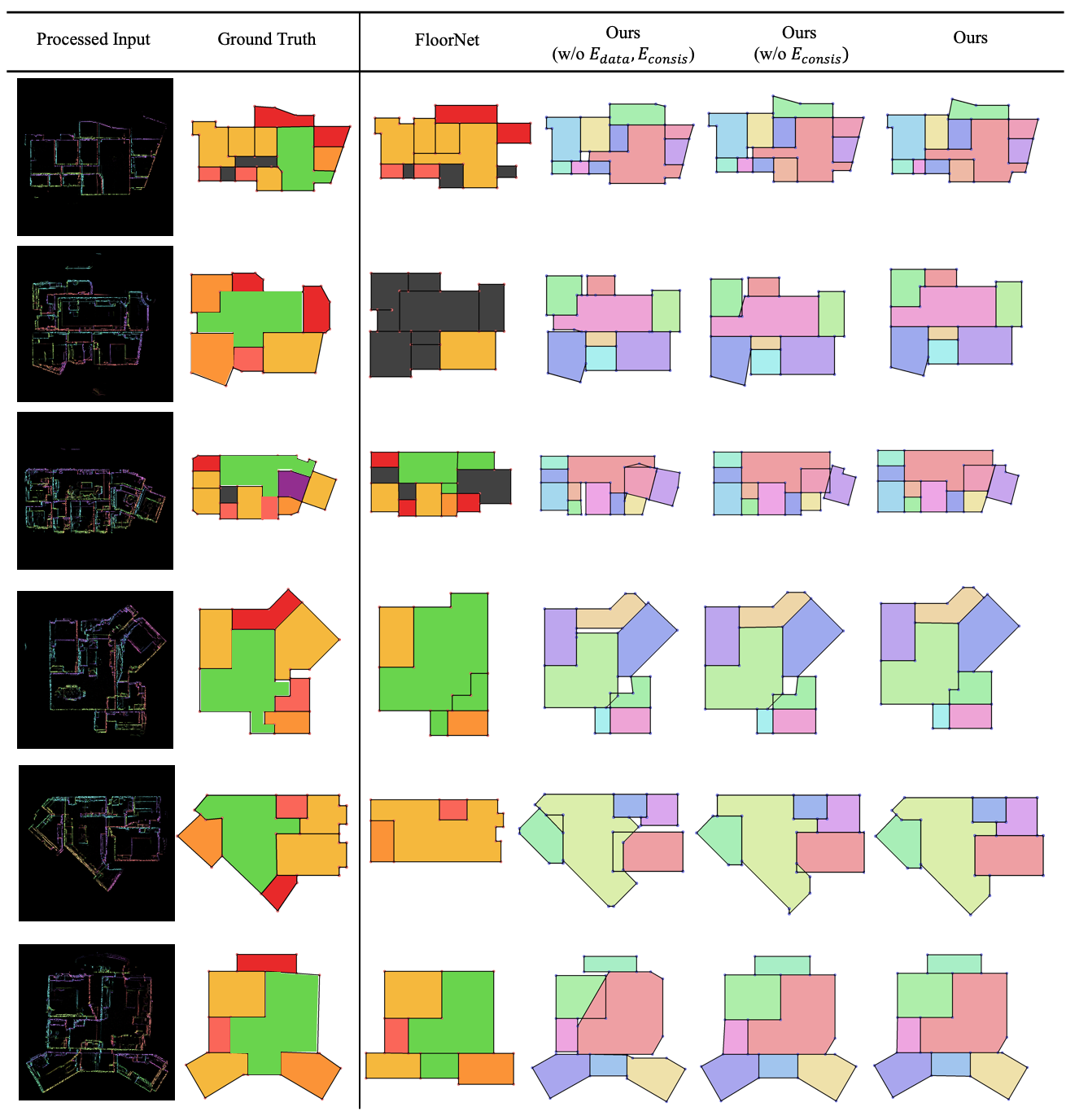}
    \caption{Qualitative comparisons against FloorNet~\cite{liu2018floornet} and the variants of our approach.
    We select hard non-Manhattan examples here to illustrate the reconstruction challenges in our dataset. 
    For reconstructions by Floor-SP variants, room colors are determined by corresponding room segments from Mask R-CNN. For the ground-truth and the FloorNet, colors are based on the room types.
    }
    \label{fig:qualitative}
\vspace{-1em}
\end{figure*}

\begin{table*}[h!]
\caption{The main quantitative evaluation results. The colors \textcolor{cyan}{cyan}, \textcolor{orange}{orange}, \textcolor{magenta}{magenta} represent the top three entries.}
  \centering
  \begin{tabular}{lcccccccc}
    \toprule
  \multirow{2}{*}{Method} & \multicolumn{2}{c}{Corner} & \multicolumn{2}{c}{Edge} & \multicolumn{2}{c}{Room}  &
  \multicolumn{2}{c}{Room++} \\
  \cmidrule(lr){2-3}\cmidrule(lr){4-5}\cmidrule(lr){6-7}\cmidrule(lr){8-9}
  & Prec. & Recall & Prec. & Recall & Prec. & Recall & Prec. & Recall\\
  \midrule
  FloorNet~\cite{liu2018floornet} & \textcolor{orange}{95.0} & 76.6 & \textcolor{cyan}{94.8} & 76.8 &  81.2 & 72.1 & 42.3 & 37.5\\
  \midrule
  Ours (w/o $E_{data}, E_{consis}$) & 84.4 & 80.4 & 82.3 & 79.8 & 75.1 & 61.3 & 23.3 & 22.0\\
  \midrule
  Ours (w/o $E_{consis}$) & 93.9 & \textcolor{orange}{82.3} & 89.2 & \textcolor{orange}{81.2} & \textcolor{magenta}{83.8} & \textcolor{magenta}{81.7} & \textcolor{magenta}{49.4} & \textcolor{magenta}{48.5}\\
  \midrule
  Ours (1st-round coordinate descent) & \textcolor{magenta}{94.6} & \textcolor{cyan}{82.8} & \textcolor{magenta}{89.4} & \textcolor{cyan}{81.7} & \textcolor{orange}{83.9} & \textcolor{orange}{81.8} & \textcolor{orange}{49.5} & \textcolor{orange}{48.7} \\
  \midrule
  Ours (2nd-round coordinate descent) & \textcolor{cyan}{95.1} & \textcolor{magenta}{82.2} & \textcolor{orange}{90.2} & \textcolor{magenta}{81.1} & \textcolor{cyan}{84.7} & \textcolor{cyan}{83.0} & \textcolor{cyan}{51.4} & \textcolor{cyan}{50.4} \\
  \bottomrule
  \end{tabular}
\vspace{-5pt}
\label{table:comparison}
\end{table*}


\subsection{Qualitative evaluations}
\label{sec:exp:qualitative}

Fig.~\ref{fig:qualitative} compares Floor-SP against the current state-of-the-art 
FloorNet~\cite{liu2018floornet} and the variants of our system.
FloorNet follows a bottom-up process, where it first detects corners then uses Integer Programming to find their valid connections. FloorNet suffers from three failure modes: 1) Missing rooms due to missing corners in the first corner detection step; 2) Extraneous rooms coming from extraneous corner detections; and 3) Broken non-Manhattan structures, which becomes challenging due to the excessive amount of search space in Integer Programming.

The right three columns show the variants of proposed Floor-SP. The left does not have the consistency term and replaces the DNN-based data term by the ad-hoc cost functions in the prior work~\cite{cabral2014piecewise}.
Our overall formulation guarantees a room reconstruction at each detected room segment, producing reasonable results.
%
%
On adding our DNN-based data term $E_{data}$ (middle),
per-room structure improves significantly.
However, inconsistencies at the room boundaries are often noticeable. 
Lastly, with the addition of the consistency term (right), we see clean floorplan structures with consistent shared room boundaries. 

\begin{figure}[!ht]
    \centering
    \includegraphics[width=\columnwidth]{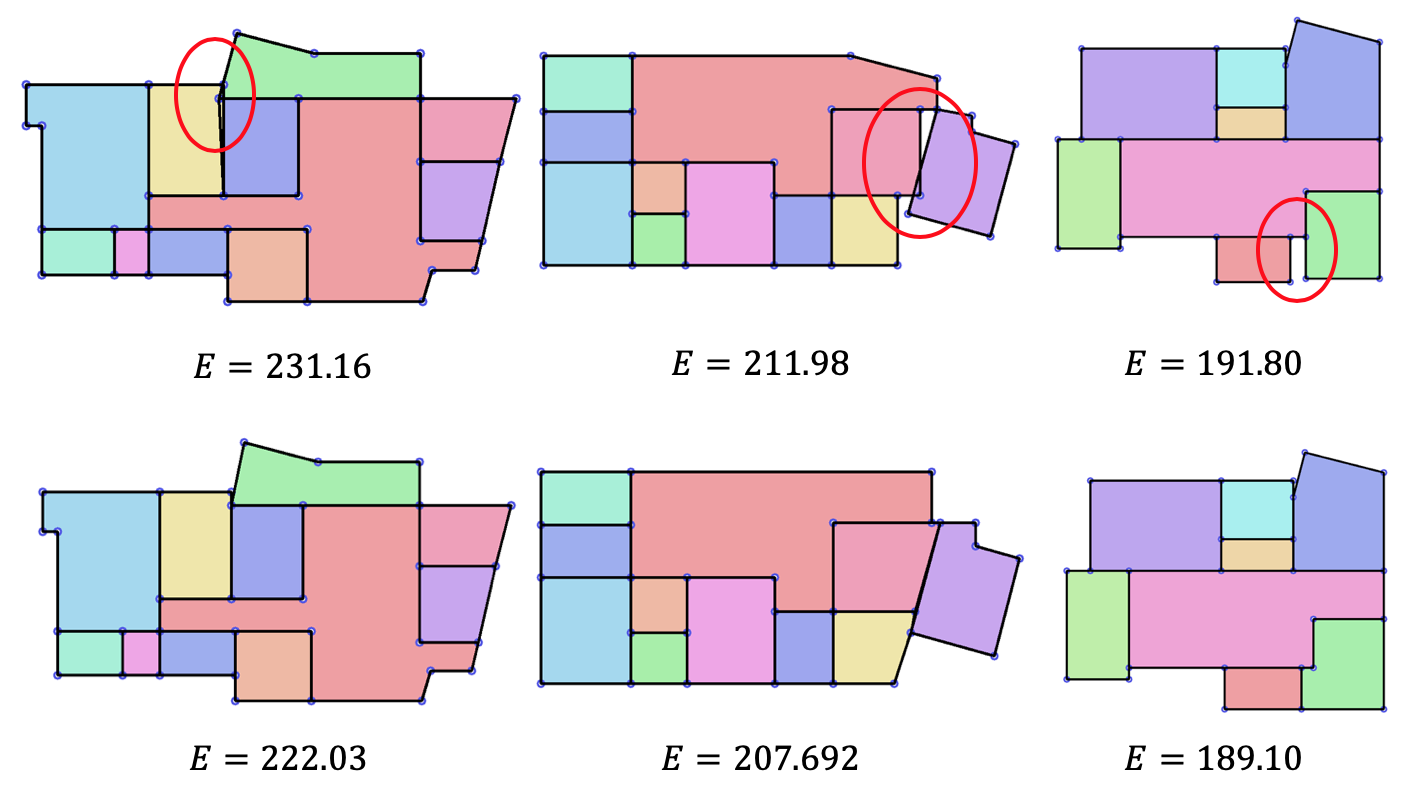}
    \caption{Multiple rounds of the coordinate descent fix mistakes at challenging floorplan structure. The top row shows the results after the first round of the coordinate descent optimization, and the bottom shows the results after the second round. We also show the total amount of energy after each round.
    %
    Corresponding ground-truth annotations are found in Fig.~\ref{fig:qualitative}.
    }
    \label{fig:energy-compare}
\end{figure}

Fig.~\ref{fig:energy-compare} illustrates the effect of room-wise coordinate descent over multiple rounds. Red ovals indicate challenging structure causing room overlaps or holes, which are resolved after the second round of optimization.



\subsection{Quantitative evaluations}


We follow FloorNet~\cite{liu2018floornet} and define the following four metrics for the quantitative evaluations:

\mysubsubsection{Corner precision/recall} We declare that a corner is successfully reconstructed if there is a ground-truth room corner within 10 pixels. When multiple corners are detected around a single ground-truth corner, we only take the closest one as correct and treat the others as false-positives.


\mysubsubsection{Edge precision/recall} We declare that an edge of a graph is successfully reconstructed if its two end-points pass the corner test described above and the corresponding edge belongs to the ground-truth.

\mysubsubsection{Room precision/recall} We declare that a room is successfully reconstructed if 1) it does not overlap with any other room, and 2) there exists a room in the ground-truth with intersection-over-union (IOU) score more than 0.7. Note that this metric does not consider the positioning and sharing of corners and edges.

%

\mysubsubsection{Room++ precision/recall} We declare that a room is successfully reconstructed in this metric, if the room is connected (i.e., sharing edges) to the correct set of successfully reconstructed rooms as in the ground-truth, besides passing the above two room conditions.
%

\mybulletend

Table~\ref{table:comparison} shows the main quantitative evaluations.
Precision metrics on low-level primitives (i.e., corners and edges) are high for FloorNet, because this task does not require high-level structural reasoning and the majority of the corners are easy ones (e.g., Manhattan corners).
On the other hand, their recall metrics are low even for low-level primitives, because some room corners do not have enough 3D points due to occlusions where DNN based corner detection fails. Floor-SP recovers such challenging corners through the sequential room-wise optimization process.
%

On room-level metrics, Floor-SP is consistently better than FloorNet. Furthermore, the addition of the data and consistency terms improves the room-level metrics. Finally, room-wise coordinate descent adds a further boost to the performance. The quantitative results and the visualization of all 94 test examples are in the supplementary document. 



\begin{figure}[!ht]
    \centering
    \includegraphics[width=\columnwidth]{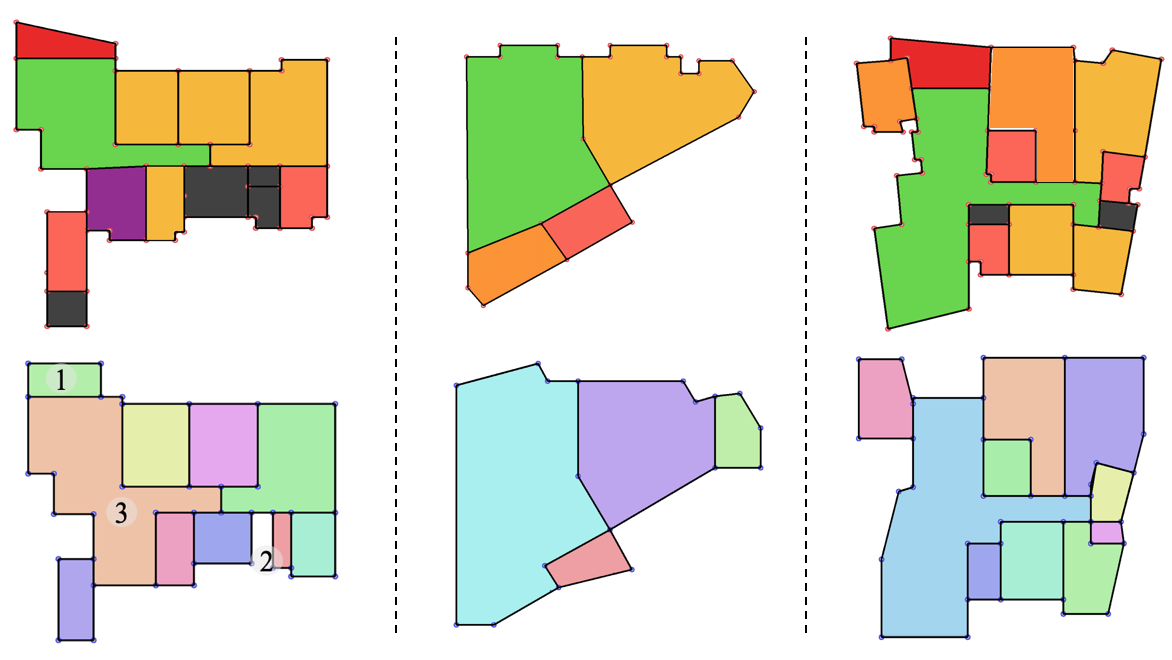}
    \caption{Typical failure modes. The top is the ground-truth annotation and the bottom is our result for each example. Our system still makes mistakes for complex scenes and challenging non-Manhattan structures.}
    \vspace{-10pt}
    \label{fig:failure-case}
\end{figure}

\subsection{Discussion}
Floor-SP produces near-perfect results for Manhattan structures. The majority of the failures are concentrated on non-Manhattan cases.
Quantitatively, our Room++ metrics are just slightly above 50. 
However, we would like to point out that our reconstructions are not terribly bad even in extremely challenging cases with poor Room++ metrics. 

Look at the first example in Fig.~\ref{fig:failure-case}. Room++ precision and recall are both 0 with our reconstruction, while the reconstruction looks fairly reasonable.
The reasons are threefold as marked by the numbers. 1) A small non-Manhattan room has wrong dominant directions in the pre-processing step, which makes it impossible for Floor-SP to recover, and fails the IOU test;
2) Small details such as concave structures are hard to keep and the room fails the IOU test; 3) The room segmentation by Mask R-CNN makes a mistake on the number of rooms for a complex case, which is again impossible to recover. 
Once a single room fails, all the adjacent rooms automatically fail in the Room++ metric, leading to the zero precision and recall in this example.

\begin{figure}[!t]
\includegraphics[width=\linewidth]{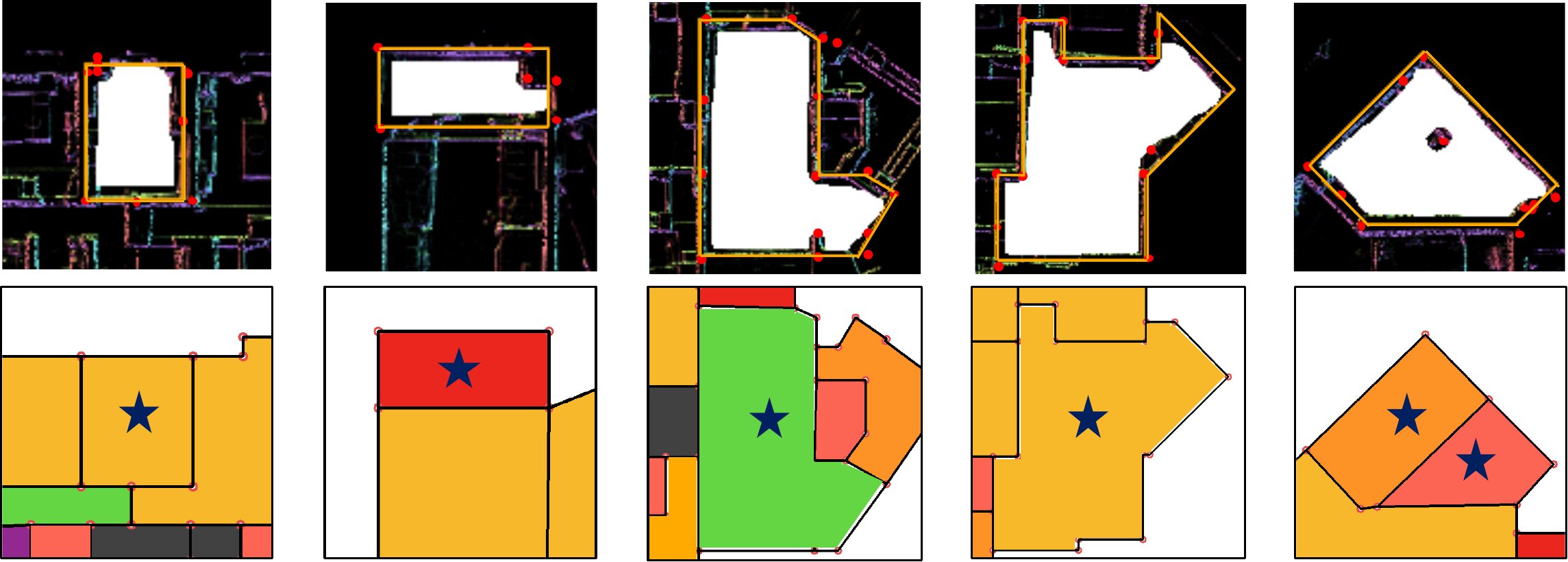}
\vspace{-0.5em}
\caption{Standard corner detection easily makes mistakes  (\textcolor{red}{red disks}). Mask R-CNN produces imprecise raster room segments (white masks) or misses an entire room (right-most example). Floor-SP uses optimization to solve for the corner placements and their connections robustly. At the top, the \textcolor{BurntOrange}{orange} polygon shows our reconstruction of a room in focus. The bottom shows the corresponding ground truth.
}
\vspace{-1em}
\label{fig:robustness}
\end{figure}

In Fig.~\ref{fig:robustness}, we further analyze the robustness of our approach. Corner detection with non-maximum suppression always produce noisy results, and room segments generated by instance segmentation network are also imprecise on details. Instead of using these primitive detections directly, Floor-SP formulates an energy minimization problem to solve for the number and placement of floorplan corners and is robust to these two types of mistakes. However, when room instance segmentation makes mistake on the number of rooms (as in the last example in Fig.~\ref{fig:robustness}), our system cannot recover but produce approximate indoor structures with wrong room separation. This mistake is also observed in the two examples in Fig.~\ref{fig:failure-case}. One future research is to recover from mistake made in room segmentation phase to produce more accurate floorplan graph.


We would like to also note that the input to our system is a single point-density/normal image from a top-down view. We have discarded the 3D information by projecting the points onto a 2D image as described in Sect.~\ref{sect:details}. We have not utilized high-resolution panorama RGB images, which are available in the dataset and could make the system more robust like FloorNet~\cite{liu2018floornet}.

We believe that this paper sets a major milestone in indoor mapping research. The proposed system produces compelling floorplan reconstruction results on production-quality challenging scenes in large quantities. We publicly share our code and data in our project website to promote further research.







\vspace{0.1cm}
\mysubsubsection{Acknowledgement}
This research is partially supported by National Science Foundation under grant IIS 1618685, NSERC Discovery Grants, NSERC Discovery Grants Accelerator Supplements, and DND/NSERC Discovery Grant Supplement. We thank Beike (\url{https://www.ke.com}) for the 3D house scans and annotations.


\clearpage

{\small
\bibliographystyle{ieee_fullname}
\bibliography{egbib}
}

\end{document}